\documentclass[fleqn,10pt]{wlscirep}
\usepackage{graphicx}
\usepackage{lineno,hyperref}
\usepackage{multirow}
\modulolinenumbers[255]
\usepackage{lmodern,textcomp}
\usepackage[caption=false]{subfig}
\usepackage{graphicx}
\usepackage[export]{adjustbox}
\usepackage{epstopdf}
\usepackage{chngcntr}
\usepackage{siunitx}
\counterwithout{equation}{section}

\title{Automatic Knee Osteoarthritis Diagnosis from Plain Radiographs: A Deep Learning-Based Approach}
\author[1,*]{Aleksei Tiulpin}
\author[1]{J\'er\^ome Thevenot}
\author[3]{Esa Rahtu}
\author[2]{Petri Lehenkari}
\author[1,4]{Simo Saarakkala}

\affil[1]{Research Unit of Medical Imaging, Physics and Technology, University of Oulu, Finland}
\affil[2]{Department of Anatomy, University of Oulu, Finland}
\affil[3]{Department of Signal Processing, Tampere University of Technology, Tampere, Finland}
\affil[4]{Department of Diagnostic Radiology, Oulu University Hospital, Oulu, Finland}
\affil[*]{aleksei.tiulpin@oulu.fi}

\begin{abstract}
Knee osteoarthritis (OA) is the most common musculoskeletal disorder. OA diagnosis is currently conducted by assessing symptoms and evaluating plain radiographs, but this process suffers from subjectivity. In this study, we present a new transparent computer-aided diagnosis method based on the Deep Siamese Convolutional Neural Network to automatically score knee OA severity according to the Kellgren-Lawrence grading scale. We trained our method using the data solely from the Multicenter Osteoarthritis Study and validated it on randomly selected 3,000 subjects (5,960 knees) from Osteoarthritis Initiative dataset. Our method yielded a quadratic Kappa coefficient of 0.83 and average multi-class accuracy of 66.71\% compared to the annotations given by a committee of clinical experts. Here, we also report a radiological OA diagnosis area under the ROC curve of 0.93. We also present attention maps -- given as a class probability distribution -- highlighting the radiological features affecting the network decision. This information makes the decision process transparent for the practitioner, which builds better trust toward automatic methods. We believe that our model is useful for clinical decision making and for OA research; therefore, we openly release our training codes and the data set created in this study.

\end{abstract}
\begin{document}
\flushbottom
\maketitle
\thispagestyle{empty}
\section{Introduction}
\subsection{Osteoarthritis: management, problems and diagnostics}
Osteoarthritis (OA) is the most common musculoskeletal disorder in the world. The literature shows that hip and knee OA are the eleventh highest global disability factor\cite{cross2014global}, thereby, causing a large economical burden to the society. It has been reported that the estimated overall costs per patient for OA treatments reach 19,000 \texteuro/year \cite{puig2015socio}. Part of these costs arise from the current clinical inability to systematically diagnose the disease at an early stage, when it might still be possible to slow down its progression or at least reduce the impact of its future disability. Because there is no effective cure for OA besides total joint replacement surgery at the advanced stage, an early diagnosis and behavioural interventions  \cite{karsdal2016disease} remain the only available options to prolong the patients' healthy years of life. Clinically, early diagnosis of OA is possible; however, currently, it requires the use of expensive magnetic resonance imaging (MRI) available only at specialised centres or in private practice. Moreover, this modality does not capture the changes in the bone architecture, which might indicate the earliest OA progression  \cite{finnila2017association}.

The current gold standard for diagnosing OA, besides the always required routine clinical examination of the symptomatic joint, is X-ray imaging (plain radiography), which is safe, cost-efficient and widely available. Despite these advantages, it is well known that plain radiography is insensitive when attempting to detect early OA changes. This can be explained by several facts: first, a hallmark of OA and the best measure of its progression is the degeneration and wear of the articular cartilage --  a tissue that cannot be directly seen in plain radiography; second, although the evaluation of the changes in the joint should be a three-dimensional (3D) problem, the imaging modality uses only two-dimensional (2D) sum projection; and finally, the interpretation of the resulting image requires a significantly experienced practitioner. Eventually, the cartilage degeneration and wear are indirectly estimated by the assessment of joint-space narrowing and bony changes, that is, osteophytes and subchondral sclerosis  \cite{kellgren1957radiological}. For these reasons, an early OA diagnosis is difficult in clinical practice.

Apart from the aforementioned limitations of plain radiography, OA diagnosis is also highly dependent on the subjectivity of the practitioner due to the absence of a precisely defined grading system. The commonly used Kellgren-Lawrence (KL) grading scale \cite{kellgren1957radiological} is semi-quantitative and suffers from ambiguity, which is reflected in the high number of disagreements between the readers (quadratic Kappa 0.56 \cite{gossec2008comparative},  0.66 \cite{sheehy2015validity}, 0.67 \cite{culvenor2015defining}). Such ambiguity   makes an early OA diagnosis challenging, thereby affecting millions of people worldwide. Such ambiguity makes an early OA diagnosis challenging, hence affecting millions of people worldwide. However, we believe that a computer-aided diagnosis (CADx) could be used as an objective tool to support clinicians in their decision.

Second, the diagnostic accuracy of these methods already reach human levels and could even outperform human experts in the future; thus, eventually the patients will obtain more reliable diagnoses. Third, supplementing the diagnostic chain with such methods, radiologists and other clinical experts can focus less on routine tasks such as image grading and focus more on incidental findings \cite{drew2013invisible}. For all the aforementioned reasons, we believe that knee OA diagnosis from plain radiographs could be significantly improved by using CADx machine learning-based methods together with clinical evaluation.

Starting in 1989, automatic knee OA diagnosis has a long history \cite{dacree1989automatic}. Although the amount of data used in these studies was previously limited to the hundreds of cases collected at a single hospital \cite{shamir2010assessment, woloszynski2012dissimilarity,shamir2009early,thomson2015automated}, research groups nowadays use thousands of cases in their analysis pipelines \cite{antony2016quantifying, antony2017automatic}. For example, Antony et al. released a study involving the combined dataset from the Osteoarthritis Initiative (OAI) and Multicenter Osteoarthritis Study (MOST) cohort baselines \cite{antony2017automatic}. The total number of images used for testing was 2,200 – approximately 1,100 subjects. The authors pointed out that the accuracy when using the combined dataset was higher than when the MOST dataset was used solely for training or the OAI dataset for testing. The method used in these studies was based on deep learning — a machine learning approach based on the automatic learning of the relevant features found in the data.

Deep learning (DL), and in particular convolutional neural networks (CNN), has recently shown ground-breaking results in a variety of general image recognition  \cite{simonyan2014very,he2016deep} and CADx tasks \cite{esteva2017dermatologist,cheng2016computer}. These powerful models already can reach human-level performance in CADx, which clearly indicates the possibility for using them in clinical practice in the near future. CNN automatically learns relevant image representations to produce a specific output, for example, diagnosis, bounding box, segmentation mask and so forth. The main disadvantages of these models, however, are that they require large datasets to be trained and the decision process is often considered a ``black box'', thus being difficult to interpret. We believe that in clinical practice, the transparency of the decision made by any automatic tool is crucial both for the practitioner and validation of the method, which is always a prerequisite before clinical use. This is one of the main reasons why the use of automatic decision support tools in clinical practice is still quite limited. Although an objective and systematic data assessment is a huge benefit in diagnostics, an understanding of what each decision was based on is another key component of decision support tools. In an optimal decision support tool, all the decisions should be transparent so that they can be checked for errors and interpreted by the clinician. While access to publicly available databases can address the training data size requirements, the development of the approaches that provide transparency of a DL-based model is still an ongoing process \cite{selvaraju2016grad,montavon2017explaining}.

Yet another critical issue related to machine learning and DL in particular is overfitting \cite{lever2016points}. This eventually results in the model's inability to perform well on the new data not seen during training. Overfitting usually occurs because of the high complexity of the model (number of parameters) \cite{lever2016points}; thus, especially in DL, where the number of parameters in the model is very high, different regularisation techniques are applied: reduction of the model's complexity, dropout, weight decay and data augmentation \cite{srivastava2014dropout,he2016deep,simonyan2014very}. To control for overfitting during training, the data are usually split into training, validation and test sets, where the training set is used for optimising the model's hyperparameters, the validation set controls for overfitting and the test set remains unseen until the model is trained. This data split is used to estimate the final generalisation error. 

All the mentioned difficulties related to overfitting are highly relevant to CADx systems' development. Such data-driven systems should eventually be robust, generalisable and able to analyse new clinical data coming from various sources, for example, at other hospitals other than where the training data were acquired. Thus, it is extremely important to validate the trained model on a test set that is completely different from the training one. Thereby, combining different medical datasets into one can be considered a limitation \cite{antony2017automatic}, if both the training and test samples are drawn from such combined data. Ideally, the independent test data should always be left out so that the generalisation measures, for example, accuracy, area under the ROC curve (AUC), mean squared error (MSE) and so forth will not be biased and will reflect the real model's performance. Furthermore, in the case of knee radiography, the variability in the images comes not only from the knee joints, but also from the imaging settings and data acquisition set-up, which could vary significantly from one dataset to another. For example, a patient imaged in a different hospital might have a different X-ray image due to these reasons. The eventual ``perfect'' prediction model should be robust enough to produce a similar output for different data acquisition settings.

\begin{figure}[!ht]
\centering
\includegraphics[width=\textwidth]{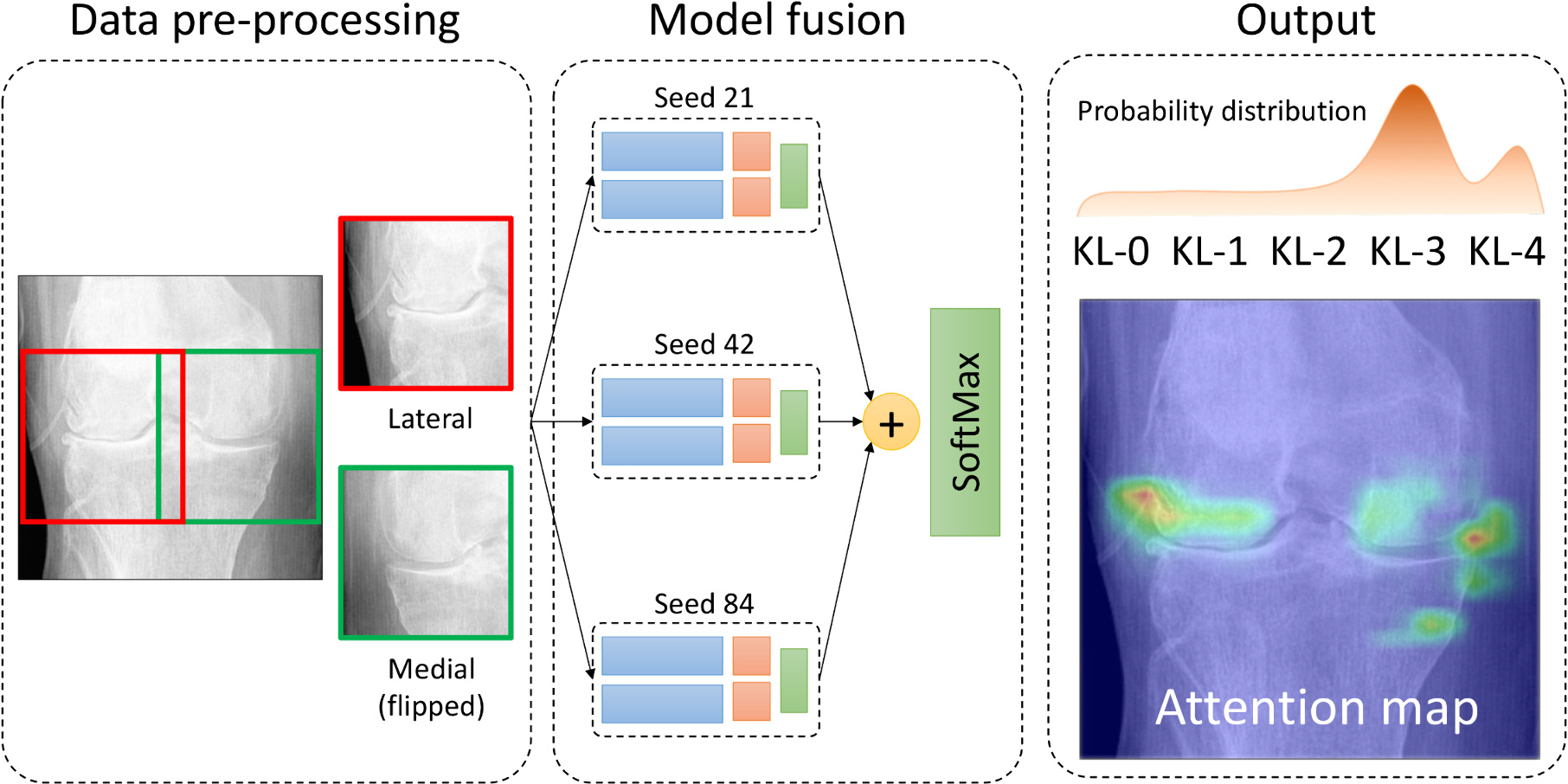}
\caption{Proposed classification pipeline. Here, we perform the knee joint area localisation, train three models using different random seeds and eventually fuse together the predictions. After this, we use the softmax layer to normalise the probability distribution and predict the resulting KL grade probability distribution  $P(y = j | \mathbf{x}), \forall j=\overline{0,4}$, where $\mathbf{x}$ is the given model input. Consequently, we also visualise the attention map, which explains the decision made by the network.}\label{fig:ens}
\end{figure}

In the present study, we demonstrate a new state-of-the-art automatic CADx method to diagnose knee OA from plain radiographs while simultaneously providing transparency in the physicians' decision-making process. Furthermore, to prove the robustness of our approach, the dataset used for training and model selection is different than the one used for the final testing. Our pipeline consists of previously published knee joint area localisation \cite{tiulpin2017} and a problem-specific CNN to grade the knee images according to the KL scale. Because the KL scale is very ambiguous and some level of uncertainty is always present in the clinical diagnosis of knee OA, our model predicts a probability distribution of the KL grades for the given image while also highlighting relevant radiological features  by generating class-discriminating ``attention maps''. Our method is schematically illustrated in Figure \ref{fig:ens}. We believe that clinically, the attention map and KL grade distribution together are highly relevant; thus, the presented approach has a clear potential to complement the OA diagnostic chain and make radiographic knee OA grading more objective.

\subsection{Novelties of this work}
In this study, we focus not only on providing new state-of-the art classification performance, but also on developing an efficient neural network architecture that learns highly relevant disease features compared to the baseline -- the fine-tuned ResNet-34 that is pre-trained on the ImageNet dataset  \cite{he2016deep}, which is motivated by the transfer learning approach \cite{esteva2017dermatologist}. Additionally, we present class-discriminating attention maps, utilising gradient-weighted class activation maps (GradCAM) \cite{selvaraju2016grad}that can be used for supplementary diagnostic information. To summarise, our study has the following novelties:

\begin{enumerate}
\item We present new state-of-the art results in automated knee OA diagnostics from plain radiographs outperforming the existing approaches \cite{antony2017automatic}.
\item We keep transparency in the decision process by providing the attention maps that show the areas of interest that contributed to the network's decision. 
\item We show a new approach for utilising Siamese deep neural networks for medical images with symmetry, which significantly reduces the number of learnable parameters, thus making model more robust and less sensitive to noise.
\item We publicly release a standardised dataset for knee X-ray OA diagnosis algorithms. 
\item Finally, we show that our method learns transferable image representation by performing an evaluation on a dataset that was not used during the training.  
\end{enumerate}

\section{Materials and Methods}
\subsection{Data}

We acquired the data from two public datasets: MOST and OAI. We used the entire MOST cohort for training our models and the baseline from the OAI dataset for testing. The MOST cohort contains the data from 3,026 subjects and their six follow-up examinations. The OAI baseline has data from 4,796 subjects. Both datasets include data from men and women aged between 50–79 years old and 45–79, respectively. The images in both cohorts were graded according to the aforementioned semi-quantitative KL scale  \cite{kellgren1957radiological}, which has five categories: KL-0 (no OA changes), KL-1 (Doubtful OA), KL-2 (Early OA changes), KL-3 (Moderate OA) and KL-4 (End-stage OA). It should be emphasised here that in this study, we did not use the OAI dataset as training material, which contrasts the previous studies by Antony \textit{et.al.} \cite{antony2016quantifying,antony2017automatic}. Our dataset consisted only from MOST images taken in \ang{5}, \ang{10} and \ang{15} beam angles. Our preliminary experiments showed that combining the projections taken from different angles creates more variability within the data.

We trained all our models for the right knee and used the horizontally flipped left ones to increase the dataset's size. From the MOST cohort, we excluded the images with implants to avoid any disturbances in the data distribution. A detailed description of the data is given in Table \ref{tab:datasets}.

\begin{table}
\centering
\caption{Description of datasets used in this study. The numbers provided in the tables indicate the number of knees used in each group. The validation set consisted of 1,502 subjects and 2,957 knee joint images. Subjects in the train and tests sets were different. The training data were used to train the models, and the validation data were used to tune the hyperparameters and evaluate overfitting. The testing set was used to estimate the final performance and contained the images from 3,000 subjects.}\label{tab:datasets}

\begin{tabular}{c c c c c c c c}
\noalign{\smallskip}
\hline
\noalign{\smallskip}
Group & Dataset &Images & KL-0 & KL-1 & KL-2 & KL-3 & KL-4 \\
\noalign{\smallskip}
\hline
\hline
\noalign{\smallskip}
Train & MOST & 18,376 & 7,492 & 3,067 & 3,060 & 3,311 & 1,446\\ 
Validation & OAI & 2,957 & 1,114 & 511& 808 & 435 & 89\\
Test & OAI & 5960 & 2,348 & 1,062 & 1,562 & 792 & 196 \\  
\hline
\noalign{\smallskip}
\end{tabular}

\end{table}

\subsection{Knee joint area localisation and side selection}\label{sec:preproc}
Using a previously developed approach \cite{tiulpin2017}, we annotated the knee joint areas so that we could use them eventually as an input for our classification system. Using this method, we selected regions of size  $140\times140$ mm according to the provided metadata: ImagerPixelSpacing DICOM tag. We performed a data augmentation and eventually used centre-cropping to obtain a region of size $130\times130$ mm.

Because we used the pre-trained model ResNet34 as a baseline, we rescaled the obtained crop to $224\times 224$ pixels. For our own model, we used the following processing strategy: First, we re-scaled the $130\times130$ mm regions to a size of $300\times 300$ pixels and then cropped two squared patches $S\times S$ pixels with the vertical offset of  $K$ pixels. Second, one patch was cropped from the lateral side and the other one from the medial side. The left top corner X-coordinate of the lateral and medial patches were 0 and $S-K$, respectively. The patch from the medial side was horizontally flipped to employ knee joint symmetry and learn the same features for both sides of the joint. The parameters $S$ and $K$ were found by optimising the score on the validation set. 

\subsection{Network architecture}
\begin{figure}[!ht]
\centering
     \subfloat[]{%
       \includegraphics[width=0.42\textwidth]{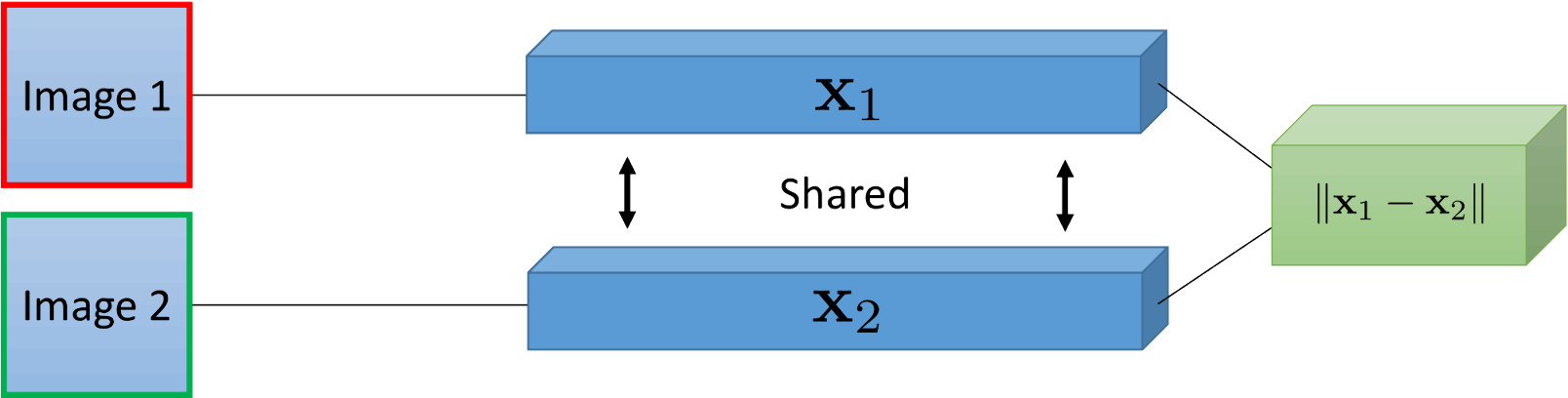}
       }
	\hfill
     \subfloat[]{%
       \includegraphics[width=0.42\textwidth]{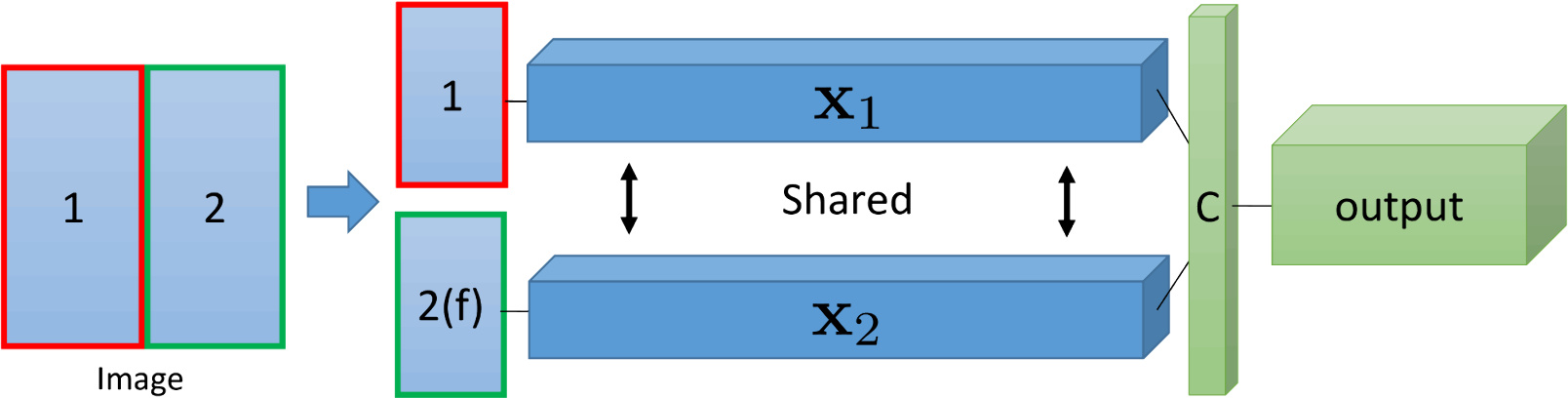}\label{fig:siam:symmetry}
       }

     \caption{Comparison between common Siamese network and our approach, in which we utilise image symmetry. Light blue rectangles denote the images. In part (a), we show a classic Siamese network application that learns a discriminative image similarity function. In this case, images are fed to the network, and the Euclidean distance is computed afterwards. In part (b), we show a symmetrical image consisting of two parts, which are the inputs for our model. 2(f) indicates the horizontal flipping of the second part. Dark blue boxes denote the shared network branches. The green box labelled as C indicates the concatenation of the outputs from the two network branches.}\label{fig:siamese}
\end{figure}

Our approach is based on the Deep Siamese CNN architecture. The original application of this architecture was to learn a similarity metric between pairs of images \cite{chopra2005learning}. Usually, the whole network consists of two branches, where each one corresponds to each input image. In our approach, we did not train our model to compare image pairs; rather, we used the symmetry in the image, which allowed the architecture to learn identical weights for every image side. The conceptual difference between our method and the traditional application of a Siamese network is illustrated in Figure \ref{fig:siamese}.

\begin{figure}
\centering
       \includegraphics[width=\textwidth]{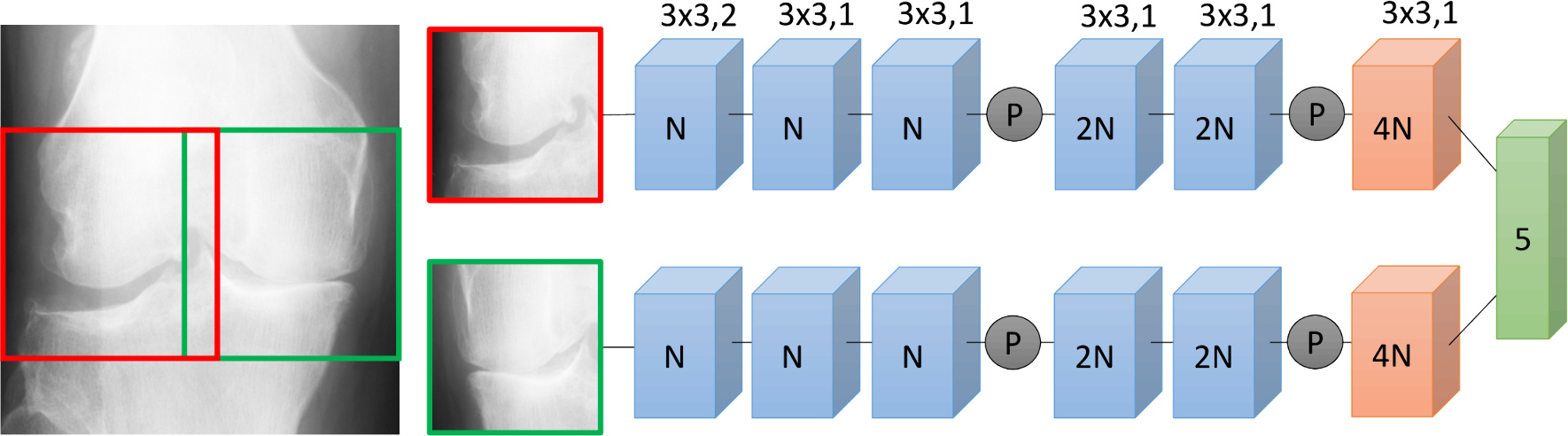}
		\caption{Schematic representation of the proposed Siamese network's architecture. First, we took the patches from the lateral and medial sides of the knee joint, horizontally flipping the latter. These patches were the inputs of the two network branches, which consisted of the following blocks having the shared weights (parameters). Blue blocks denote convolution (Conv), batch normalisation (BN) and rectified linear unit (ReLU) layers. Grey circles indicate a max-pooling  $2\times2$. Light-red blocks consist of Conv-BN-ReLU layers followed by the global average pooling. The final green block is a softmax layer (classifier) taking a concatenation of the two network branches outputs and predicting KL grade probability distribution over five grades. The numbers inside the Conv-BN-ReLU blocks indicate the number of feature maps (convolutional filters), and the numbers on top of them indicate their parameters  $K\times K,S$, where $K$ is a filter size and $S$ is the convolution stride.}\label{fig:arch}
\end{figure}

The proposed network consists of two branches, each having convolution, max-pooling, rectified linear unit (ReLU), batch normalisation  \cite{ioffe2015batch}, global average pooling and concatenation layers, as described in Figure  \ref{fig:arch}. As mentioned above, we learned the same convolutional and batch normalisation weights for both branches because we employed the symmetrical structure of the knee joint. Here, we did not consider the knee joint to be symmetrical, but only the features that are learned, that is, we assumed that learned edge-detection features are not different for the lateral and medial sides. Eventually, we concatenated the outputs from the lateral and medial branches and used a final fully connected layer to make the prediction.

Our proposed application of a Siamese network can also be related to the recent developments in neural networks for fine-grained classification: recurrent attention networks  \cite{zhao2017diversified, mnih2014recurrent}. In this approach, the sequence of image locations is adaptively selected for further prediction. However, here, we did not select the relevant image regions adaptively because the image structure was known. In particular, we constrained the network attention to only two regions on the knee joint’s sides (see Figure \ref{fig:arch}). Such an approach is close to the real radiological diagnosis and grading conducted by a human evaluator. By the definition found in the KL system, knee joint tibial and femoral corners, as well as the joint space, are used in the grading. Thus, we explicitly mapped the relevant attention zones to the network input and took only these into account when making the decision. Moreover, because knee joints are relatively symmetrical, we learned the same weights for the medial and lateral sides by flipping the medial side horizontally (illustrated in Figures \ref{fig:ens}, \ref{fig:siam:symmetry} and \ref{fig:arch}). This allows to drastically reduce the number of learnable parameters and constrain the network to learn only the relevant features used also by human evaluators.

The design of the individual network branches in our model was motivated by the CNN design approach presented by Simonyan \textit{et al.}\cite{simonyan2014very}. Therefore, we used a stack of $3\times 3$ convolutional filters instead of the larger ones. However, in contrast to \cite{simonyan2014very}, we did not employ zero padding for $3\times 3$ filters, which does not change the input size and used to build very deep networks. Because the problem of knee OA diagnosis from plain radiographs is fine-grained \cite{antony2017automatic}, we wanted to keep a strong signal from the first layers; thus, we designed the network branches using only five convolutional layers, where each one reduced the input size. For this, we also used max-pooling $2\times 2$ and a convolution stride (see, Figure \ref{fig:arch}) of 2 in the first layer. In our architecture, we used a global average pooling at the end of each branch to keep the higher resolution, helping the system learn more local features, because we believe that they are relevant to OA diagnosis.

\subsection{Models ensemble and inference}\label{sec:inference}
Considering the real-life situation when an OA diagnosis is done by a practitioner, the diagnosis can be wrong due to the limitations mentioned in the introduction. However, the OAI and MOST datasets include data graded by several readers to ensure their reliable KL grading. In this work, we imitated this process by training three separate models initialised with  different random seeds. Subsequently, we picked the models that best performed on the validation set, summed their predictions and propagated them through the softmax layer. Eventually, the class probability of the KL grade $j$ for given image $\mathbf{x}$ was inferred as follows: 

\begin{equation}\label{eq:inference}
P(y=j\mid \mathbf {x} )=\frac {\exp\left[\sum_{m=1}^M \hat P_m(y=j\mid \mathbf {x})\right]}{\exp\left[\sum_{k=1}^{K}\sum_{m=1}^M \hat P_m(y=k\mid \mathbf {x})\right]},
\end{equation}
where the number of models in the ensemble $M=3$, number of classes $K=5$, and $ \hat P_m(y=j\mid \mathbf {x})$ an unnormalised probability distribution (individual network output before the softmax layer). The described approach is illustrated in Figure  \ref{fig:ens}.

\subsection{Ensembled GradCAM for Siamese networks attention visualization}\label{sec:gradcam}
The problem with automatic KL grading is that it is fine-grained, and moreover, only several thousands of cases are available to train the model on. Therefore, the models with large number of parameters may overfit to the background noise or react to image artefacts instead of paying attention to the disease-relevant features.

Here, we would like to highlight the importance of understanding these automatic methods, such as deep neural networks, because they automatically learn the relevant features to produce the target label. As mentioned above, the learnt features may not be relevant to the disease; thus, it is important to examine them and the region where the network is looking. In this work, we utilised a state-of-the art approach, GradCAM \cite{selvaraju2016grad}, which allowed us to obtain class-discriminating activation maps. Following the original methodology \cite{selvaraju2016grad}, here, we describe a modification of the method, which can be applied to Siamese networks and their ensembles.

Let us denote $A_i^{l_k}$ a $k^{th}$ activation map of size $X\times Y$ from the layer of interest $l$ belonging to the branch $i$ and $y_c$ -- the output of the network. The attention map of the branch $i$ with respect to the class $c$ was computed as follows:
\begin{equation}
\overline A_{c_i} = \text{ReLU}\left(\sum_k w_{i_k}^cA_i^{l_k}\right),
\end{equation}

where $\text{ReLU}(\cdot)$ is a mapping, substituting the negative input values with 0, and $w_{i_k}^c$ is the class-specific weights, which are found by computing the global average pooled gradient from the following network layer $l+1$:
\begin{equation}\label{eq:weights}
w_{i_k}^c = \frac{1}{XY}\sum_x\sum_y\frac{\partial y_c}{\partial A_i^{l_k}}(x,y).
\end{equation}

These gradients are obtained by guided backpropagation of the prediction $c$. In the original GradCAM approach, the incoming gradients were averaged manually for each activation map  $A_i^{l_k}$, while in the presented above architecture, those are computed automatically, since we chose $l$ to be a penultimate layer of the network. Thus, the averaging in the equation \ref{eq:weights} can be omitted because $X=Y=1$. Since our model has two branches, the final attention map with respect to the class $c$ is a pair $\left \{\overline A_{c_1},\overline A_{c_2}\right\}$.

In the previous section, we presented our ensembling approach of the Siamese branches. In the case of the attention maps, we perform exactly the same procedure as we do for a single model; however, in this case, we add the attention maps for lateral and medial branches, respectively: 

\begin{equation}
\left\{\overline A_{c_1}, \overline A_{c_2}\right\} = \left\{ \sum_{k=1}^{M}\overline A_{c_1}^{m},\sum_{k=1}^{M} \overline A_{c_2}^{m}\right\}
\end{equation}

After obtaining a pair $\left\{\overline A_{c_1}, \overline A_{c_2}\right\}$,  we horizontally flip the map corresponding to the medial side and project both maps back to the original image. Finally, we use a min-max normalisation to equalise the final attention map.

\subsection{Implementation details}
\paragraph{Dataset distribution and pre-processing}
We used our previously developed knee joint area localisation method  \cite{tiulpin2017}. Mislocalised knee joints (1.5\%) in the test sets were manually re-annotated to increase their size. In total, our training, validation and test sets included 18,376; 2,957 and 5,960 images, respectively. Further details regarding the data distribution are presented in Table \ref{tab:datasets}. It should be mentioned that the left knee images were horizontally flipped to be similar to the right ones. The parameters $K=100, S=128$ were selected to generate the input pair of patches out of the knee joint image (see section \ref{sec:preproc}).	

We used 16-bit DICOM files and converted them to an 8-bit resolution, with a preceding global contrast normalisation and histogram truncation between the $5^{th}$ and $99^{th}$ percentiles. This step was performed to reduce the impact of noise. For testing the images, we non-linearly changed the contrast of the over- and underexposed images using gamma correction.

During the training we balanced the data using oversampling and eventual bootstrapping. We applied random rotation, contrast, brightness, jitter and gamma correction data augmentations to the bootstrapped samples. Further parameters of the data augmentation are supplied with our model training scripts. After the data augmentation, we cropped the images $130\times 130$ mm. All augmentations were performed in a random order on the fly. 

\paragraph{Model training}

We used PyTorch\cite{pytorch} and 4$\times$Nvidia GTX1080 cards with 8 GB memory for each of the experiments. In the study, we compared three different settings with a fixed random seed of 42. First, we evaluated our network architecture with different values of the parameter, and for each of these settings, we compared the shared weights implementation versus the case when the network branches are not shared (see Figure \ref{fig:arch}). Secondly, we implemented the previously published method by Antony \textit{et al.}\cite{antony2017automatic}. Finally, motivated by the transfer learning approach \cite{esteva2017dermatologist}, we fine-tuned a ResNet-34 network \cite{he2016deep} pre-trained on the ImageNet dataset. To train all the analysed models, we used a mini-batch size of 64.

In the experiments done on our architecture, we used Adam optimizer \cite{kingma2014adam} and a cross-entropy loss function. We optimised the network stochastically, giving it a learning rate of  $1e-2$ for 50,000 iterations evaluating every 500 iterations. To combat overfitting, we used L2-norm regularisation (weight decay) with the coefficient  $1e-4$ and a dropout \cite{srivastava2014dropout} of 0.2. The dropout was inserted after the concatenation of the branches (see, Figure \ref{fig:arch}). The regularisation parameters were optimised based on the validation set loss. Eventually, we used this procedure three times with random seeds 21, 42 and 84 and selected the best model snapshots for these three settings  ($M=3$ models). We found the aforementioned configuration the most promising from the model selection. Complete simultaneous training of the whole model ensemble (seeds 21, 42 and 84) for 50,000 iterations per one training epoch took roughly six hours.

To train the network by Antony \textit{et al.} \cite{antony2017automatic}, we implemented the architecture and the custom loss function proposed in the article. For this experiment, we trained the network using stochastic gradient descent and Nesterov momentum, dropout, weight decay and the aforementioned data augmentations. In total, we trained the model for 250 000 iterations with the epoch size of 500 iterations. 

To train the fine-tuned ResNet-34, we used Adam method with a learning rate of  $1e-3$ for 14,300 iterations, having the epoch size of 300 iterations. We inserted a dropout of 0.5 before the final linear classifier and used a weight decay of  $1e-4$. We provide further details of the training procedure and model selection of all mentioned models in the supplementary information.

\section{Experiments and Results}

\subsection{Knee osteoarthritis diagnosis}
As described, we trained multiple networks and used the Kappa metric on the validation set to select the baseline for the final comparison with our best-performing model. Fixing the seed to 42, we found our shared weights model  $N=64$ and a fine-tuned ResNet-34 network yielding the best validation set performance. 

After the model selection, we trained an ensemble of the networks, as described in section \ref{sec:inference}, for seeds 21 and 84, and we used the already obtained model for the random seed value of 42. For each of the seeds, we selected the snapshot of the network that yielded the best validation set performance. Selected snapshots were corresponded to 32,000 (seed 21), 48,500 (seed 42) and 48,000 (seed 84) iterations. 

\begin{figure}
\centering
     \subfloat[Confusion matrix]{
       \includegraphics[width=0.4\textwidth]{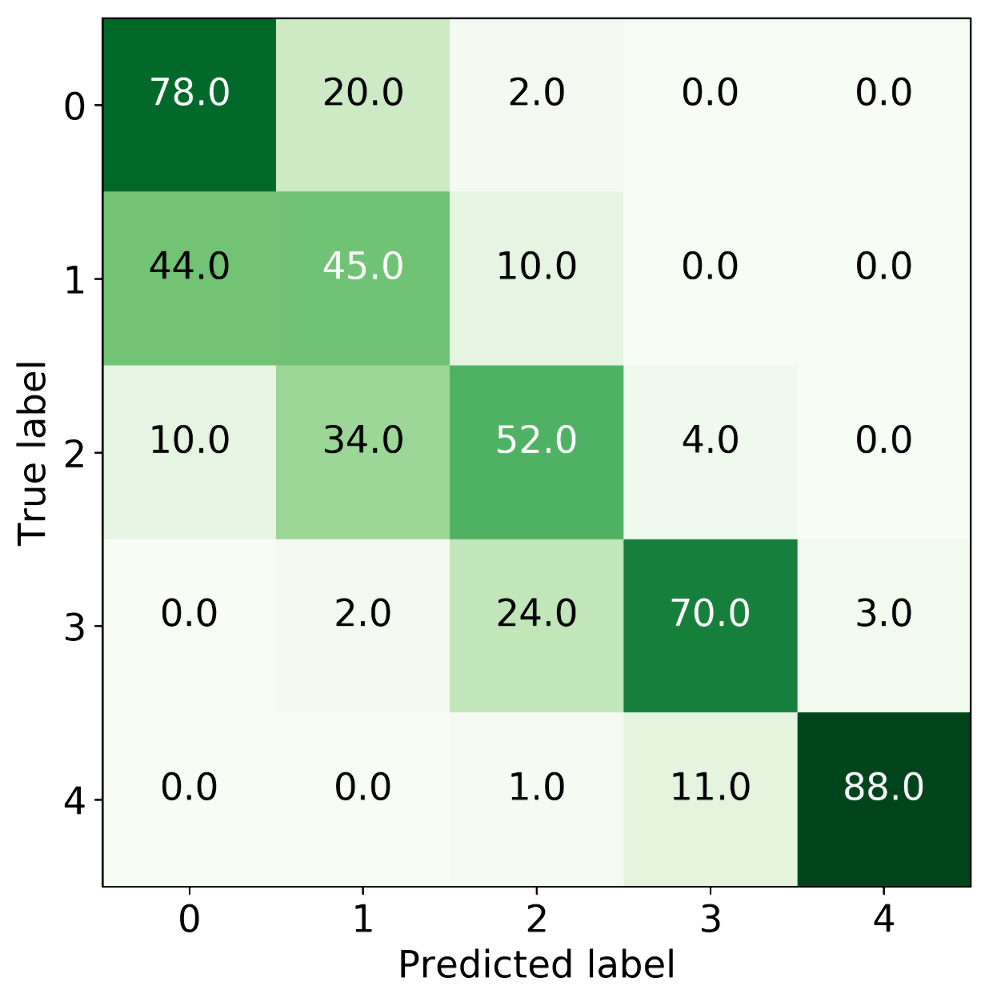}
        }
        \hfill
     \subfloat[ROC-curve]{
       \includegraphics[width=0.4\textwidth]{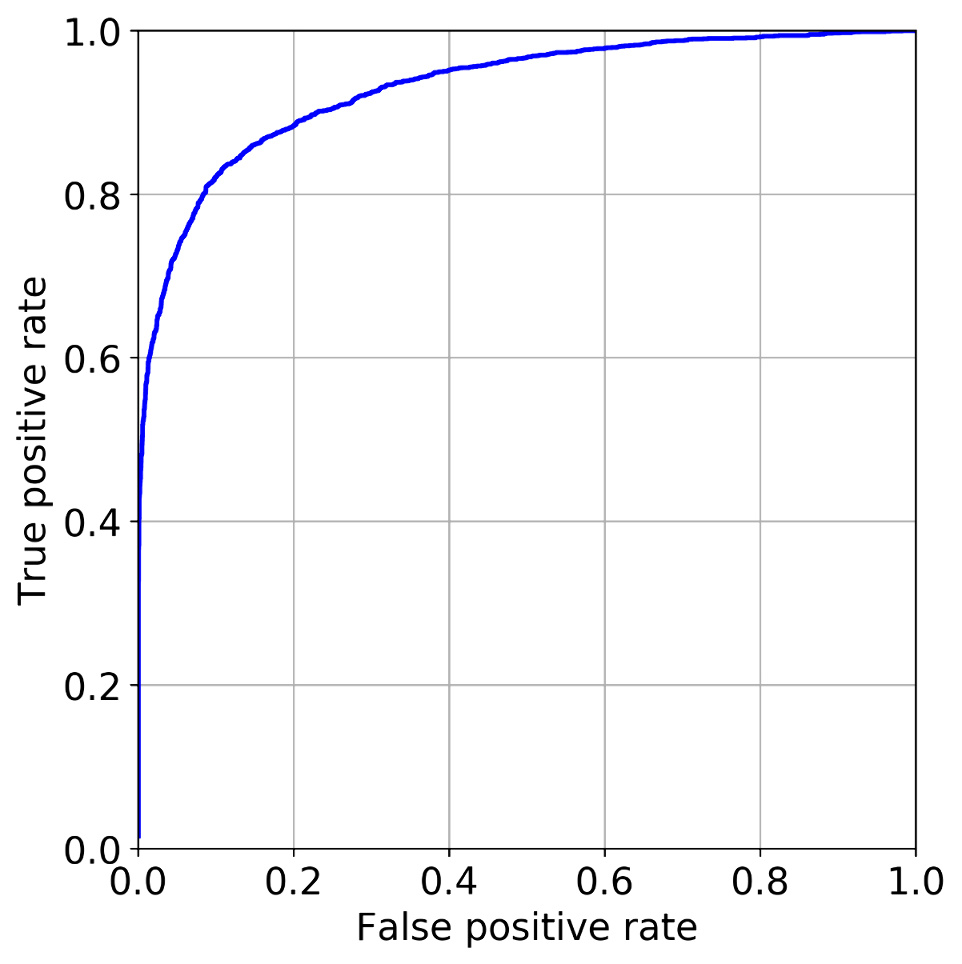}
        }

     \caption{(a) Confusion matrix of KL grading and (b) ROC curve for radiographic OA diagnosis KL $\geq2$ produced using our method. Average multi-class accuracy is 66.71\%, and AUC value is 0.93. Corresponding Kappa coefficient and MSE value are 0.83 and 0.48, respectively.}\label{fig:auc}
\end{figure}

We performed the final testing solely using the OAI dataset while the training was done using only MOST data. The final model evaluation was performed on 3,000 subjects (5,960 knee joints in total) randomly selected from the OAI dataset baseline. The average test set multi-class accuracy achieved by our method was 66.71\%. The confusion matrix is presented in Figure \ref{fig:auc}. Furthermore, in Figure \ref{fig:auc}, we show the radiographical OA diagnosis ROC-curve (KL$\geq2$). The achieved AUC of 0.93 was higher than any previously published results \cite{minciullo2016fully,thomson2015automated}. Importantly, considering the fact that OAI data were not used as a training material, these results indicate a good clinical applicability of the method.

We also compared the quadratic Kappa coefficient and MSE values because they were used previously in clinical studies \cite{gossec2008comparative, culvenor2015defining}. The Kappa metric reflects the agreement between two raters and weighs the different misclassification errors differently, for example, mistake 1 versus 2 has less impact on the score than 0 versus 4. In this case, we considered our model to be an X-ray reader (evaluator) and used its predictions to assess the agreement between its predictions and the expert annotations from the OAI dataset. Our Kappa value on the test set was 0.83, which is considered to be an excellent agreement between the raters \cite{landis1977measurement}. The classification MSE value achieved was 0.48, which is lower than previously published results \cite{antony2017automatic}. 

As mentioned before, besides our method, we also evaluated a fine-tuned ResNet-34 network because it performed similarly on the validation set. On the test set, the baseline also performed similarly to our approach in terms of MSE (value of 0.51), Kappa agreement (value of 0.83) and average multi-class accuracy (value of 67.49\%). However, the qualitative assessment showed (see, the next section) that despite having a similar performance, the fine-tuned model sometimes pays more attention to the regions that do not have relevant radiological findings. Additionally, when comparing the classification accuracy of the most clinically relevant case KL-2, our method outperformed the baseline by roughly 4\% (52\% with our method versus 48\% baseline). parameters and that the ImageNet distribution significantly differs from our data distribution. Thereby, such an extensive network has a high possibility of overfitting to the background noise and finding other possible relevant correlations present in the data, for example, bone texture  \cite{thomson2015automated}, patella intensity and so forth.

\subsection{Class-discriminating attention maps and prediction confidence}

\begin{figure}[!ht]
\subfloat[KL-2 -- ground truth]{%
      \begin{minipage}{0.27\textwidth}
        \includegraphics[width=1\textwidth,valign=t]{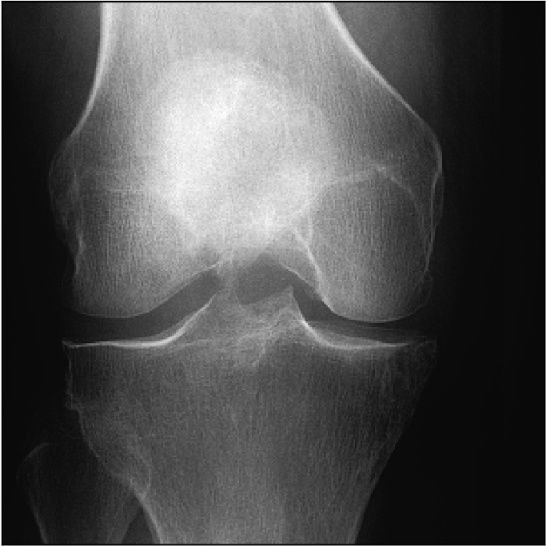}\\
        \vspace{31pt}
      \end{minipage}
       }
            \hfill
     \subfloat[KL-2 -- ResNet-34]{%
     \begin{minipage}{0.27\textwidth}
     	\includegraphics[width=1\textwidth,valign=t]{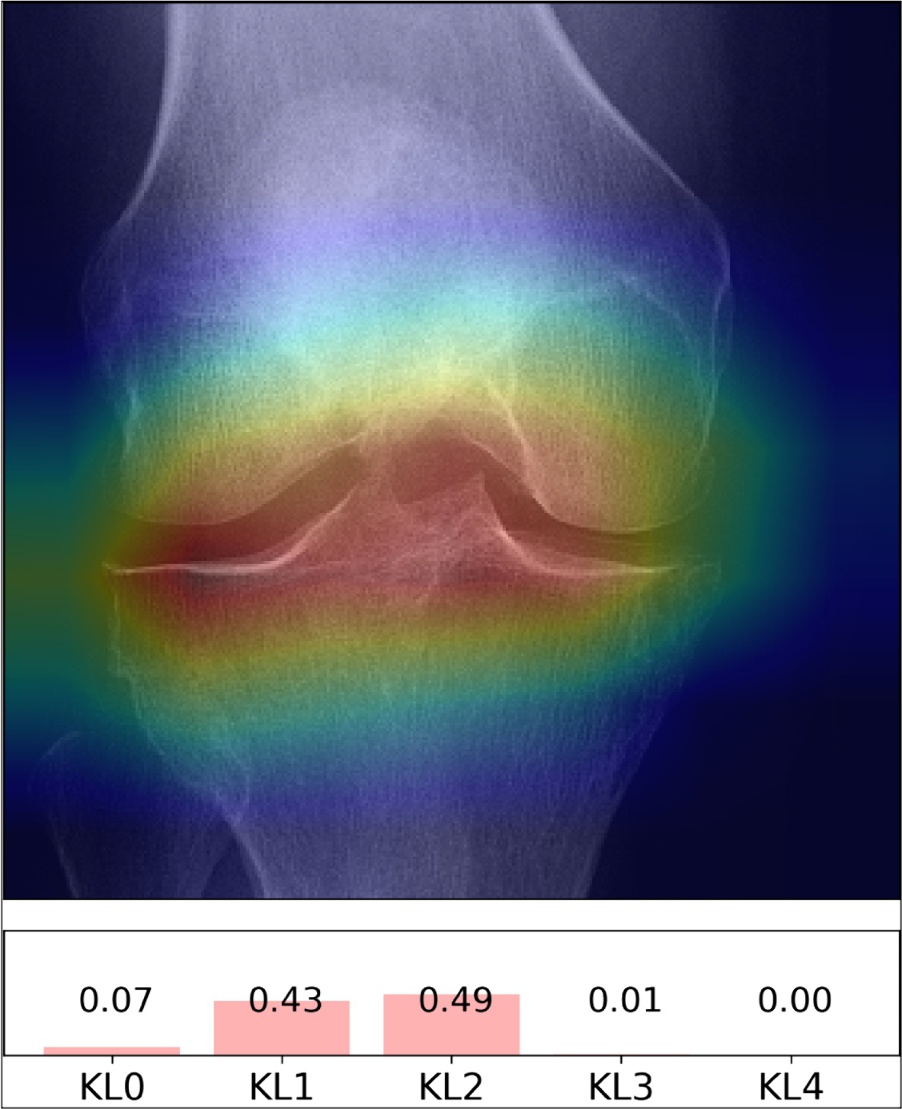}
     \end{minipage}
       }
     \hfill   
        \subfloat[KL-2 -- Our model]{%
        \begin{minipage}{0.27\textwidth}
       		\includegraphics[width=1\textwidth,valign=t]{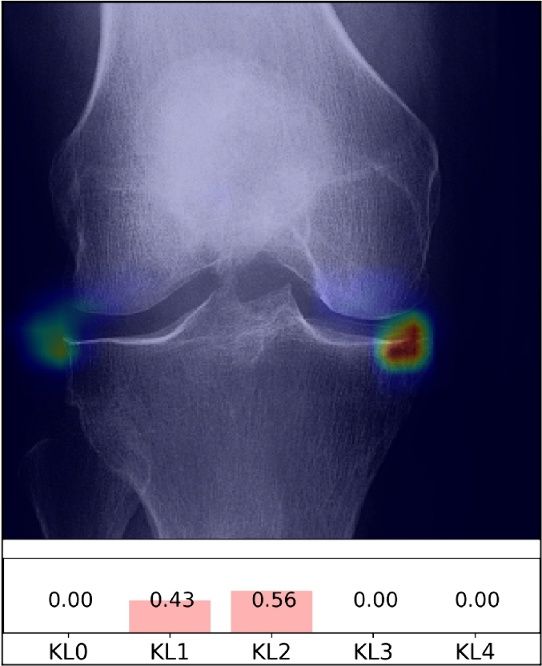}
        \end{minipage}
       }
     \caption{Comparison of the attention maps of the correctly classified examples between the baseline and our method. The original image (a) is the attention map produced from the last residual block in the baseline model (b), and the attention map produced by our model is (c). From the presented example images, the baseline can react to the background noise values or bone texture in classification. Underneath (b) and (c), we present the predicted probabilities. Attention maps show that our model reacts to the relevant radiological findings -- osteophytes -- while the baseline reacts to the joint centre. }
\label{fig:attention}
\end{figure}

We obtained the visualisations for the baseline method, ResNet-34, as described in \cite{selvaraju2016grad}, and visualisations for our network using its modification, as described in section \ref{sec:gradcam}. The pre-trained baseline model paid the strongest attention to the image regions, which do not have the findings used by the radiologists for the decision. For example, the strongest network activation comes from the centre of the knee joint, surrounding fat tissue or bone texture. In contrast, the attention map of our model clearly learns more local features that highlight the true relevant radiological findings. The most probable reason is that we imposed domain-knowledge constraints (prior anatomical knowledge) to the network’s architecture, thereby forcing it to learn only the features related to the radiographical findings, such as osteophytes, bone deformity and joint-spacing narrowing, which are all used to grade the image according to the KL scale (see, Figure \ref{fig:arch}) .

Here, we also report another clinically relevant result: the probability distribution of the KL grades over the images. This information comes inherently from the network’s architecture and can be used as another source of supplementary diagnostic information (see, Figures \ref{fig:ens} and \ref{fig:attention}). For example, if the model is not confident in the prediction, this is seen in the distributions. Further examples of the attention maps and the probabilistic outputs are also provided in the supplementary information section.

\section{Discussion}
In the present study, we demonstrated a novel approach for automatically diagnosing and grading knee OA from plain radiographs. In contrast to previous studies, our model uses specific features relevant to the disease, ones that are comparable to the ones used in clinical practices (e.g., bone shape, joint space, etc.). Furthermore, compared to the previously published approaches, our method achieves the highest multi-class classification results, despite having a different testing set: average multi-class accuracy of 66.71\%, radiographical OA diagnosis AUC of 0.93, quadratic weighted Kappa of 0.83 and MSE of 0.48. If we compare this to the average human agreements in KL grading (0.5-0.8), our method achieves a very high quadratic Kappa, indicating it can perform at a human level.

An important issue to consider in our model is that it was trained solely with the MOST dataset and tested with the OAI dataset. The main advantage of this study's design was the demonstration of the model's ability to learn relevant OA features that are transferable across different datasets. This clearly indicates that our method is robust toward different artefacts and data acquisition settings. To create a clinically applicable model, we considered multiple steps to enhance its robustness. First, we normalised the data to always have a constant region-of-interest  ($130\times130$ mm) and constrained the attention zones by taking into account only the regions of interest used by a radiologist when making the decision. Second, we included the full MOST cohort, including image data from the same subject multiple times, that is, used several follow-up examinations. This increased the training data size. Third, we included the radiographs taken \ang{5}, \ang{10} and \ang{15} X-ray beam angles, which helped to regularise the training and induced more variability to the dataset. Fourth, we used rotation, jitter, contrast and brightness data augmentation techniques, which made our training more robust. Finally, we used an ensemble of three networks trained with different random seeds, which induced less variance into the model decisions.

There were also several limitations in our study. Our validation set was selected from the OAI dataset, but an alternative approach would be to keep these data out. However, our method performed the best among the compared models from a clinical point of view: it learned local radiographical findings and yielded better classification performance of early OA cases than other approaches. Despite this, in our future studies, we will investigate the generalisability of the method across multiple datasets using larger amounts of data. Another limitation is that we reduced the image resolution to 8 bits, which could have led to the loss of fine-grained information stored in the images. It is possible that the use of the original image’s resolution with reasonable data filtering could further improve our results. Apart from this, the attention maps from the baseline method (fine-tuned ResNet-34 motivated by a transfer learning approach) had a lower resolution than the ones produced by our model -- the network from the last ReLU layer in the model was only  $7\times7$ pixels than the output of our model -- so each branch of the network had $10\times 10$ pixels output. Despite this limitation, it is still evident that the baseline method learned only those image representations that best correlated to the target variable (KL grade) while having similar performance with our method. Furthermore, in some of the misclassified images, our radiologist and orthopaedist strongly disagreed with the ground truth KL grades. Thus, it is possible that in certain limited cases, the KL grade in the OAI dataset is erroneous. Our version of the OAI dataset had images from the releases 0.C.2 and 0.E.1, while the newer releases are also available now, and these could include the corrected KL grades for those individual cases. Finally, we would like to mention that our method could be further improved by utilising a different loss-function that optimises the Kappa coefficient itself, as well as by using larger amounts of training data from different sources. Finally, the images used in this study were obtained in standardised settings, including the positioning frame. Consequently, the method cannot be directly adapted to every health care practice, and further research is needed to understand how our model, trained with MOST dataset, would perform on the images acquired without such a frame.

As mentioned above, in contrast to the baseline, our approach was capable of learning highly localised radiographical findings from knee images (see Figure \ref{fig:attention}). An important benefit of our method is the supplementary information produced by the attention maps. We believe that having this attention map in the automatic CADx systems will eventually build better trust in the clinical community regarding these the artificial intelligence based methods. Additionally, we proposed to use a probability distribution of the grades over the images to assess KL grading CADx systems (see, Figure \ref{fig:ens} and the supplementary information section). We believe that having such outputs could provide further information to the practitioner, showing that the severity of the disease is not a finite grade. By providing the probability for specific KL grades, the model mimics the decision process of the practitioner: choosing between two KL grades by considering the closest one to the medical definition. This could highly benefit inexperienced practitioners and eventually decrease their training time.

To conclude, we believe that the proposed approach has several benefits. First, it can help patients suffering from knee pain receive a faster diagnosis. Second, health care in general will benefit by reducing the costs of routine work. Although the present study focused on OA, our model possesses the ability to systematically assess a patient’s knee condition and monitor it for other conditions (e.g., follow-up of ligament surgery, assessment of joint changes after knee unloader prescription, etc.). Third, the research community will benefit from utilising our method as a tool with which to analyse large cohorts, such as OAI and MOST. To boost such research, we provide a standardised benchmark for automatic OA radiographical grading methods comparison. Here, we release a public dataset that contains bounding boxes for the MOST cohort and OAI cohort baselines used for our experiments. We also provide pre-trained models and training codes for all analysed models. Our codes and datasets are publicly released on GitHub: \url{https://github.com/lext/DeepKnee}.

\section*{Acknowledgments}

The OAI is a public-private partnership comprised of five contracts (N01-
AR-2-2258; N01-AR-2-2259; N01-AR-2- 2260; N01-AR-2-2261; N01-AR-2-2262)
funded by the National Institutes of Health, a branch of the Department of Health
and Human Services, and conducted by the OAI Study Investigators. Private
funding partners include Merck Research Laboratories; Novartis Pharmaceuticals
Corporation, GlaxoSmithKline; and Pfizer, Inc. Private sector funding for the
OAI is managed by the Foundation for the National Institutes of Health.

MOST is comprised of four cooperative grants (Felson - AG18820; Torner
 - AG18832; Lewis - AG18947; and Nevitt - AG19069) funded by the National
Institutes of Health, a branch of the Department of Health and Human Services,
and conducted by MOST study investigators. This manuscript was prepared
using MOST data and does not necessarily reflect the opinions or views of MOST
investigators.

We would like to acknowledge the strategic funding of the University of Oulu. Additionally, we want to thank Iaroslav Melekhov for the useful discussions about Siamese networks and GradCAM.

\section*{Author contributions}
A.T. originated the idea of the study, performed the experiments and wrote the manuscript. J.T. helped with the study design and wrote the manuscript. E.R. participated in the study design and provided editing at the final stage. P.L. provided clinical feedback, participated in clinical data interpretation and edited the article at the final stage. S.S. contributed at each stage of the process and supervised the project.

\section*{Additional information}
\subsection*{Competing financial interests}
The authors declare no competing financial interests.

\bibliography{mybibfile}

\section{Supplementary Information}
\subsection{Model selection process and comparison to the previous work}							
In our implementation, we used the PyTorch framework and 4$\times$NVidia GTX1080 cards. Because the MOST dataset was very imbalanced because of the low presence of higher KL grades, we used oversampling to overcome this issue in all our experiments: for each training epoch, we randomly sampled with repetitions roughly  $N_{cat}\times B$ images from each of the categories (KL 0–4), where $N_{cat}$ is the average number of training examples per category in our training set and $B$ is a bootstrap factor. From our training data we found $N_{cat}\times B=3,675$. Parameter $B=15$ was found empirically by trial and error. We found this strategy useful to prevent overfitting, especially when it is combined with data augmentation and selection of the number of batches per epoch (Table \ref{tab:benchmarking}). For data augmentation, we used brightness, contrast, rotation, gamma correction and jitter. In our experiments, we mostly used Adam’s method with a learning rate of  $1e-2$; however, to reproduce the results presented by Antony et. al. in \cite{antony2017automatic}, we used a stochastic gradient descent with Nesterov momentum and learning rate of  $1e-4$. The batch size which was used in all our experiments was empirically selected to be 64.

We systematically compared multiple models — multiple configurations of our proposed approach. First, we re-implemented the best-performing network described in the article by Antony et al. This network produces two outputs — one for classification and the other one for regression. The optimisation is done by minimising an average of mean squared error (MSE) and cross-entropy. The idea behind this loss function is to give a network information about the importance of higher (e.g., KL4) versus lower (KL0) misclassifications. In our implementation, we cropped the $300\times 300$ to $300\times 200$ pixel images and used them as the network input, as described in the manuscript. The $300\times 300$  images were obtained after the data augmentation. Due to the insufficient implementation details provided in the original paper and the differences in our training settings, we could not exactly reproduce the results; however, we found validation performance in the multi-class average accuracy and MSE that were similar to the values reported by the authors. To achieve these results, we had to use the following strategy: starting from the learning rate of $1e-4$ we were dropping 10 times it each 50,000 iterations. When the learning rate drop was less than $1e-6$, we increased it back to $1e-4$ and continued this procedure. In total we trained the network for 250,000 iterations. This was performed because of the plateau in training, and it helped to escape the achieved local minima.

Secondly, we performed a fine-tuning of a ResNet-34 network \cite{he2016deep} that was pre-trained on the ImageNet dataset. We found this model overfitting quickly so decided to evaluate it more frequently -- 300 iterations per training epoch compared to our model and the model from Antony \textit{et al.} \cite{antony2017automatic} -- 500 iterations per training epoch. In total, we trained ResNet-34 for 14, 300 iterations and had to stop the process afterwards because the validation loss started to rapidly increase. We summarise all our results in Table \ref{tab:benchmarking}. Based on the validation Kappa, we selected the fine-tuned ResNet and our model with $N=64$ for a qualitative comparison, as described in the article and the next section.

\begin{table}[]
\centering
\caption{Model selection and comparison to the other models. Here, $N$ in the own models indicates the number of filters in the first layer, as in the main text of the article, and indicates whether the weights of the network branches were shared or not. \# Batches indicates the epoch size, Kappa corresponds to the quadratic Kappa coefficient, MSE to the mean squared error and Accuracy the average multi-class accuracy. All the models were trained with a batch size of 64 samples. Column Kappa shows in bold the two best models -- our models with the starting number of filters $N=64$ and the fine-tuned ResNet-34.}
\label{tab:benchmarking}
\begin{tabular}{lcccccc}
\hline
\noalign{\vskip 1mm}    
\multicolumn{1}{c}{\textbf{Model}} & \multicolumn{1}{l}{\textbf{Learning rate}} & \textbf{\# Batches}  & \textbf{Optimizer}          & \multicolumn{1}{c}{\textbf{Kappa}} & \multicolumn{1}{c}{\textbf{MSE}} & \multicolumn{1}{c}{\textbf{Accuracy}} \\ \noalign{\vskip 1mm}  \hline\hline
Own {[}N=32{]}                      & \multirow{6}{*}{$1e-2$}           & \multirow{6}{*}{500} & \multirow{7}{*}{Adam} & 0.803                              & 0.526                            & 67.04                                 \\
Own {[}N=32, NS{]}          &                                 &                      &                                 & 0.706                              & 0.732                            & 56.40                                 \\
Own {[}N=64{]}                      &                                 &                      &                         & \textbf{0.808}                     & 0.518                            & 64.77                                 \\
Own {[}N=64, NS{]}          &                                 &                      &                                 & 0.718                              & 0.736                            & 57.81                                 \\
Own {[}N=128{]}                     &                                 &                      &                         & 0.801                              & 0.515                            & 66.35                                 \\
Own {[}N=128, NS{]}         &                                 &                      &                                 & 0.727                              & 0.705                            & 58.78                                 \\
ResNet-34                           & $1e-3$                          & 300                  &                         & \textbf{0.812}                     & 0.512                            & 67.02                                 \\
Model by Antony \textit{et al.}, 2017      & $1e-4$                          & 500                  & SGD                     & 0.770                              & 0.670                             & 59.52                                 \\ \hline
\end{tabular}
\end{table}

\subsection{Attention maps and probability distribution examples}
\begin{figure}[!ht]
	\subfloat[KL-2 -- ground truth]{%
       \includegraphics[width=0.29\textwidth,valign=t]{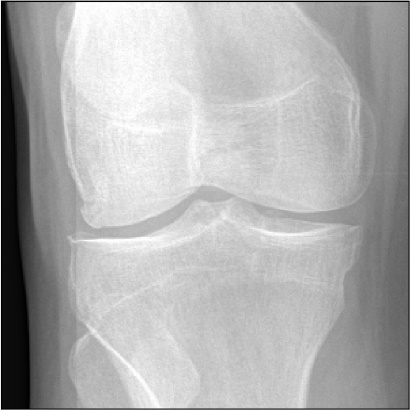}
       }
            \hfill
     \subfloat[KL-2 -- ResNet-34]{%
       \includegraphics[width=0.29\textwidth,valign=t]{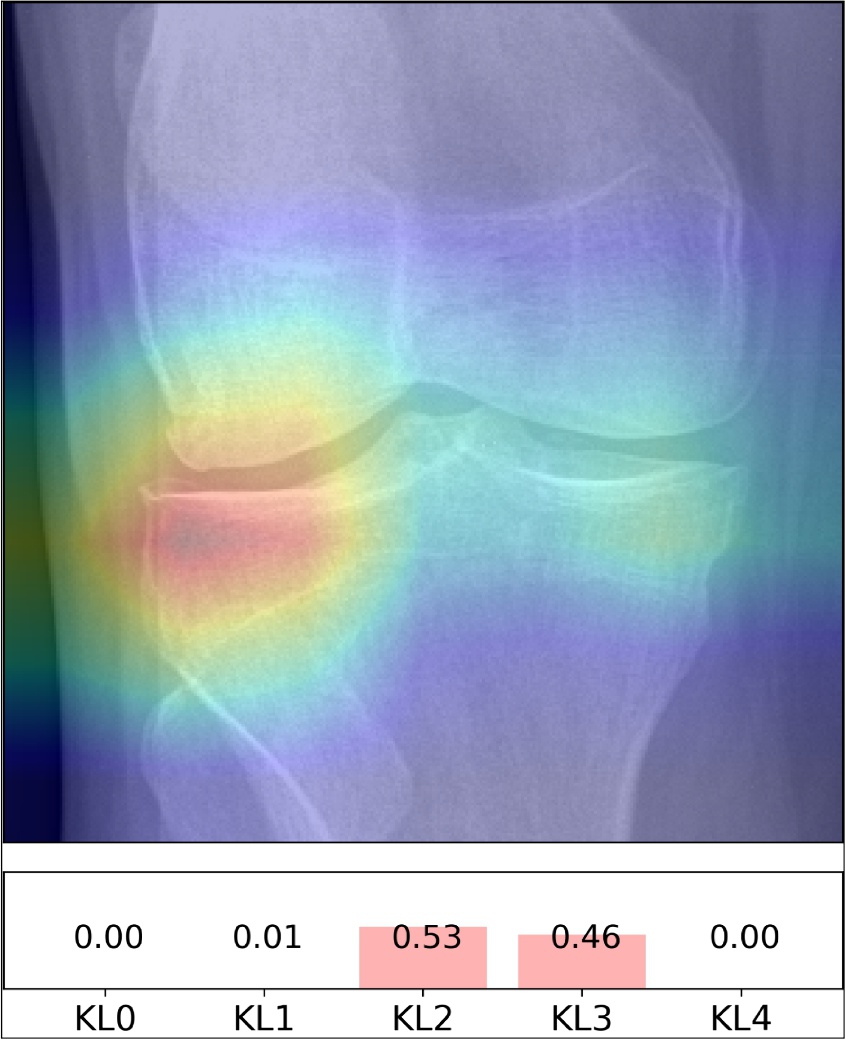}
       }
        \hfill   
     \subfloat[KL-2 -- Our model]{%
       \includegraphics[width=0.29\textwidth,valign=t]{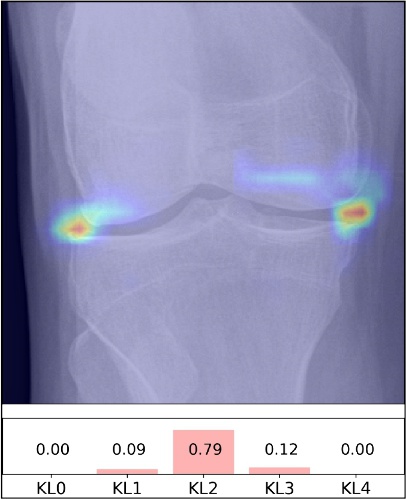}
     }
     
	\subfloat[KL-2 -- ground truth]{%
       \includegraphics[width=0.29\textwidth,valign=t]{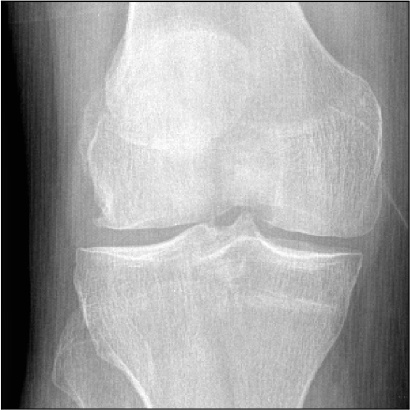}
       }
            \hfill
     \subfloat[KL-2 -- ResNet-34]{%
       \includegraphics[width=0.29\textwidth,valign=t]{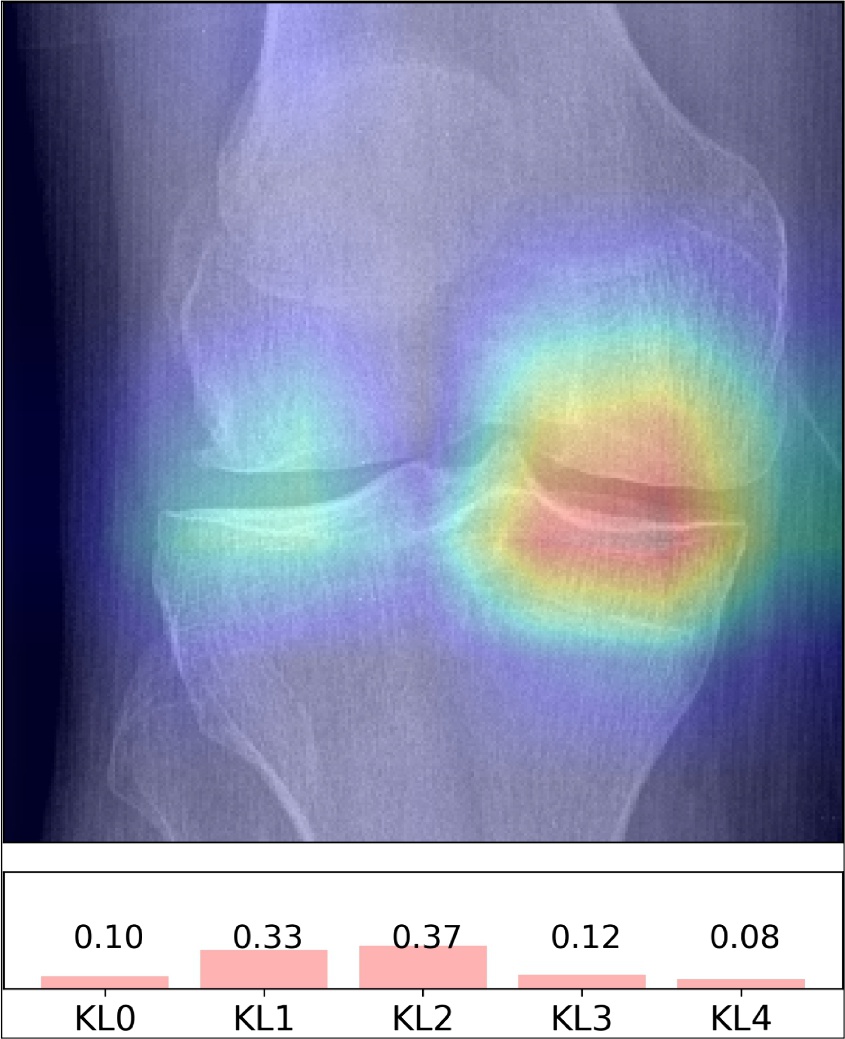}
       }
                \hfill   
     \subfloat[KL-2 -- Our model]{%
       \includegraphics[width=0.29\textwidth,valign=t]{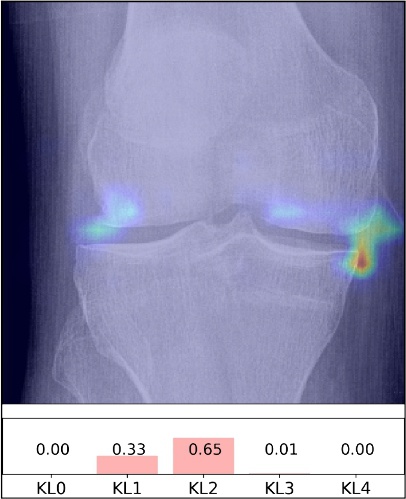}
     }

    \subfloat[KL-2 -- ground truth]{%
       \includegraphics[width=0.29\textwidth,valign=t]{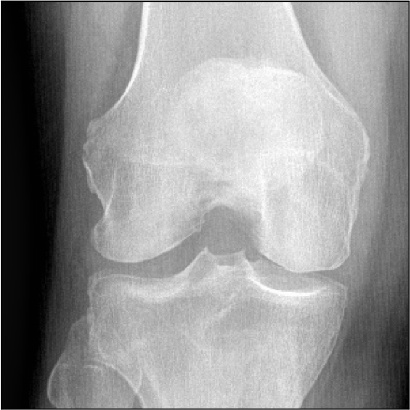}
       }
            \hfill
     \subfloat[KL-1 -- ResNet-34]{%
       \includegraphics[width=0.29\textwidth,valign=t]{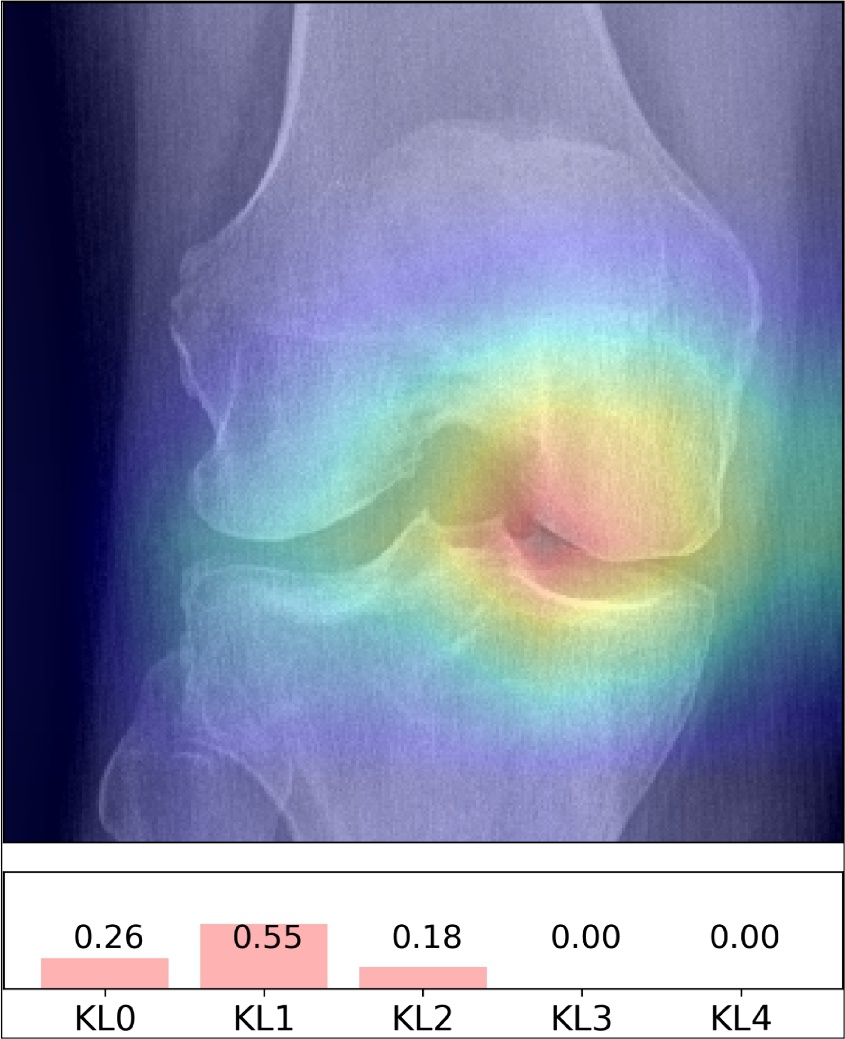}
       }
                \hfill   
     \subfloat[KL-2 -- Our model]{%
       \includegraphics[width=0.29\textwidth,valign=t]{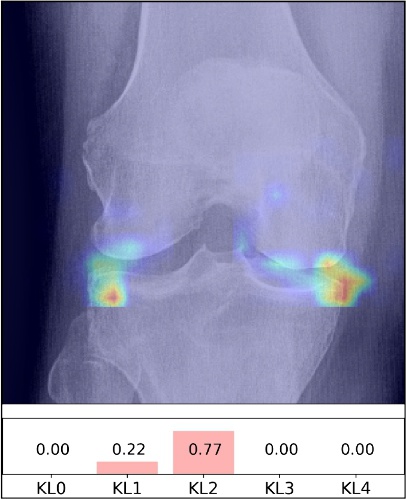}
     }     
     
     \caption{Comparison of the attention maps and output probability distributions between the baseline and our method for the clinically relevant case KL-2. The examples show that the pre-trained model is less certain than our proposed approach.}\label{fig:attention_app1}
\end{figure}

\begin{figure}[!ht]
	\subfloat[KL-3 -- ground truth]{%
       \includegraphics[width=0.29\textwidth,valign=t]{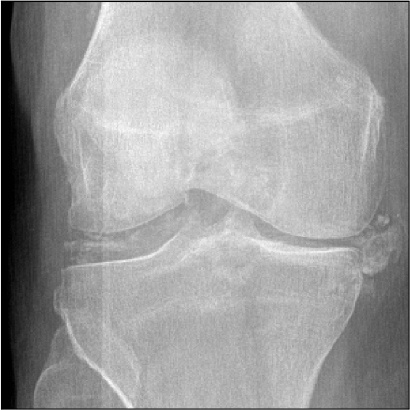}
       }
            \hfill
     \subfloat[KL-4 -- ResNet-34]{%
       \includegraphics[width=0.29\textwidth,valign=t]{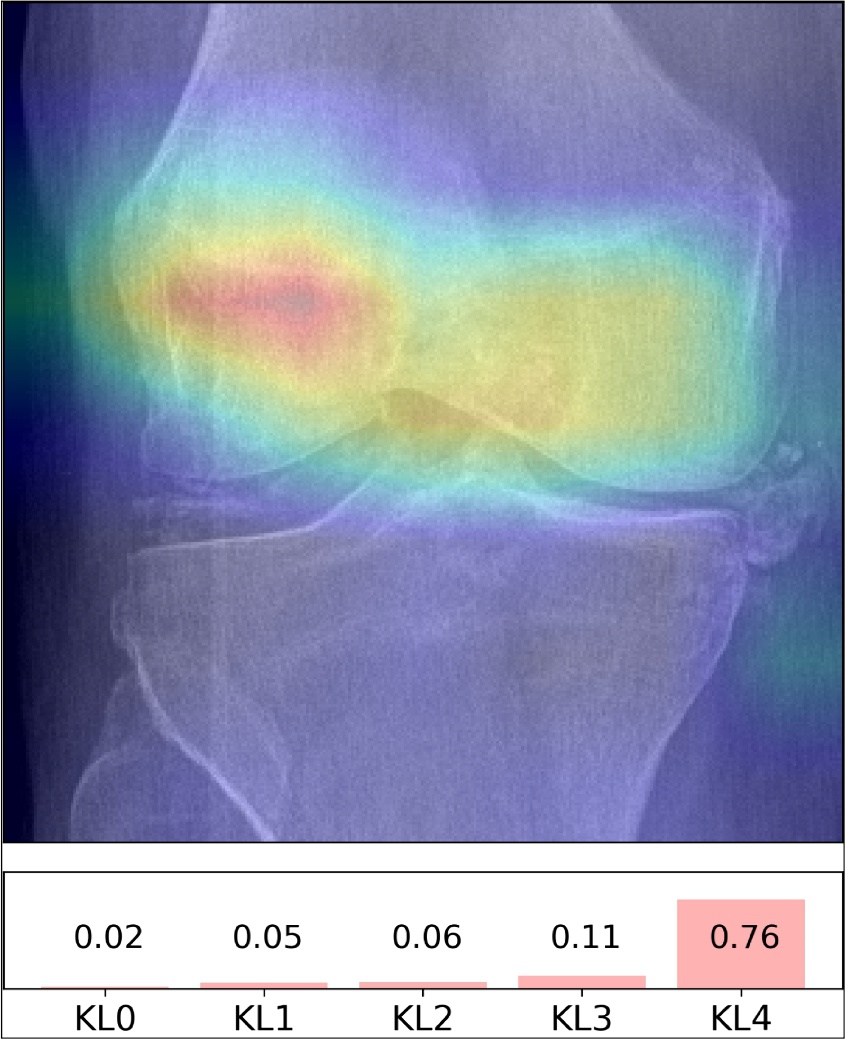}
       }
        \hfill   
     \subfloat[KL-3 -- Our model]{%
       \includegraphics[width=0.29\textwidth,valign=t]{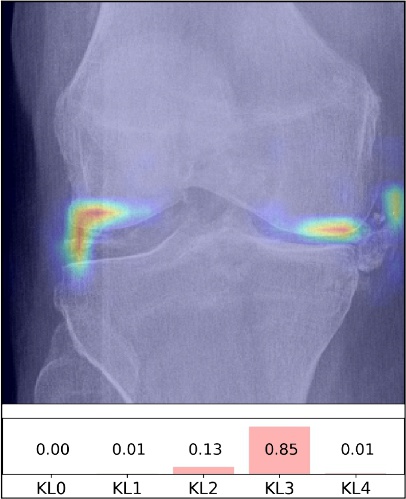}
     }
     
	\subfloat[KL-3 -- ground truth]{%
       \includegraphics[width=0.29\textwidth,valign=t]{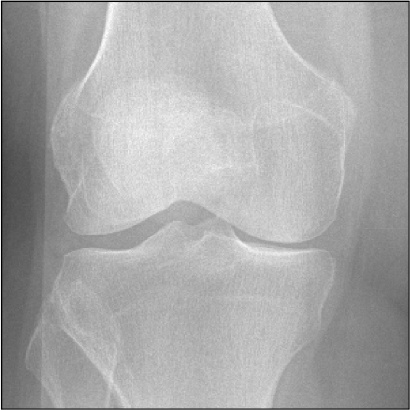}
       }
            \hfill
     \subfloat[KL-2 -- ResNet-34]{%
       \includegraphics[width=0.29\textwidth,valign=t]{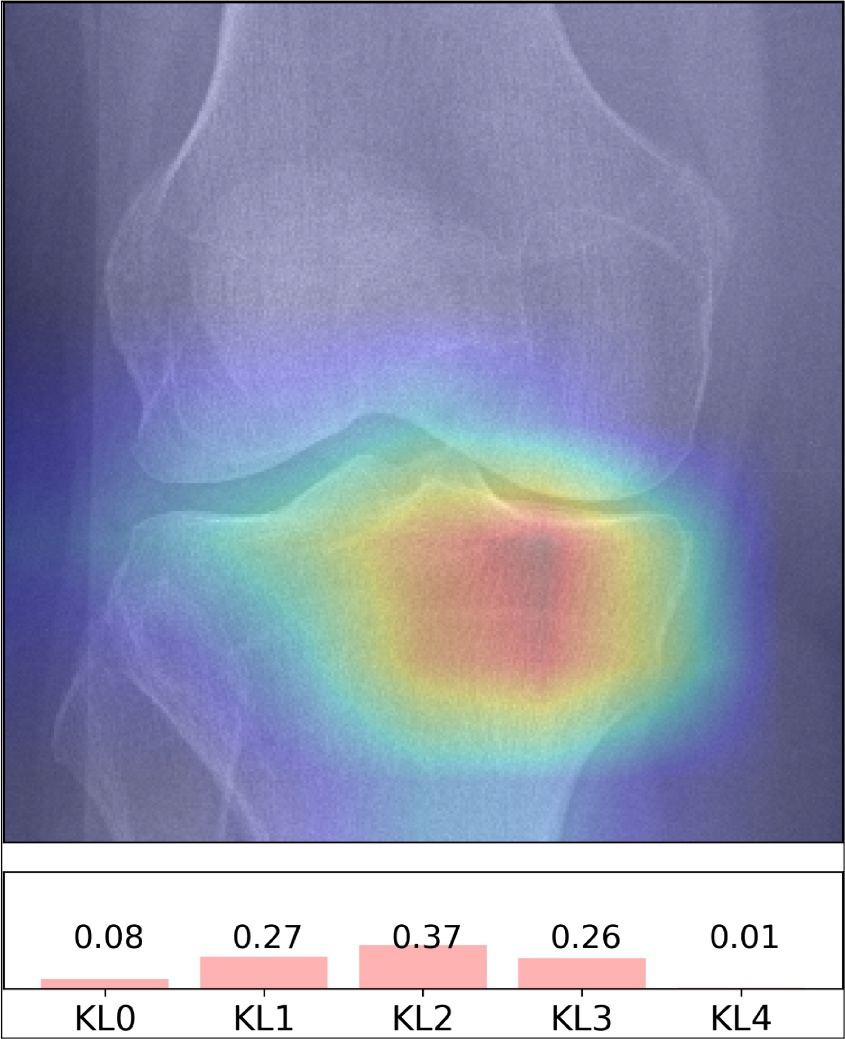}
       }
                \hfill   
     \subfloat[KL-2 -- Our model]{%
       \includegraphics[width=0.29\textwidth,valign=t]{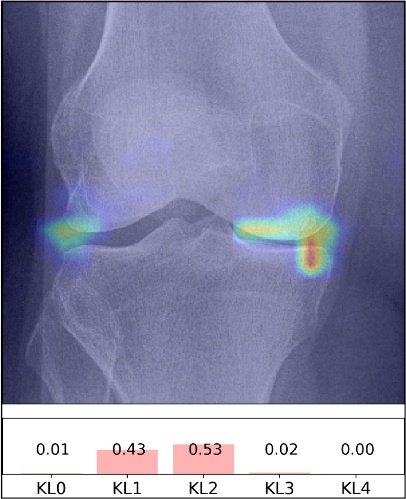}
     }

    \subfloat[KL-3 -- ground truth]{%
       \includegraphics[width=0.29\textwidth,valign=t]{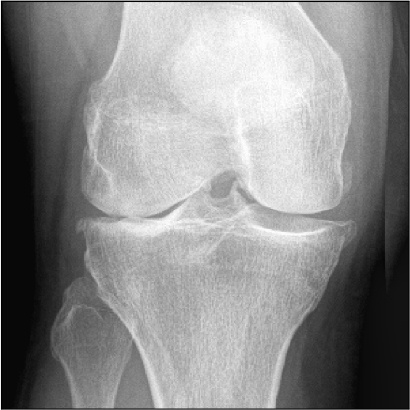}
       }
            \hfill
     \subfloat[KL-3 -- ResNet-34]{%
       \includegraphics[width=0.29\textwidth,valign=t]{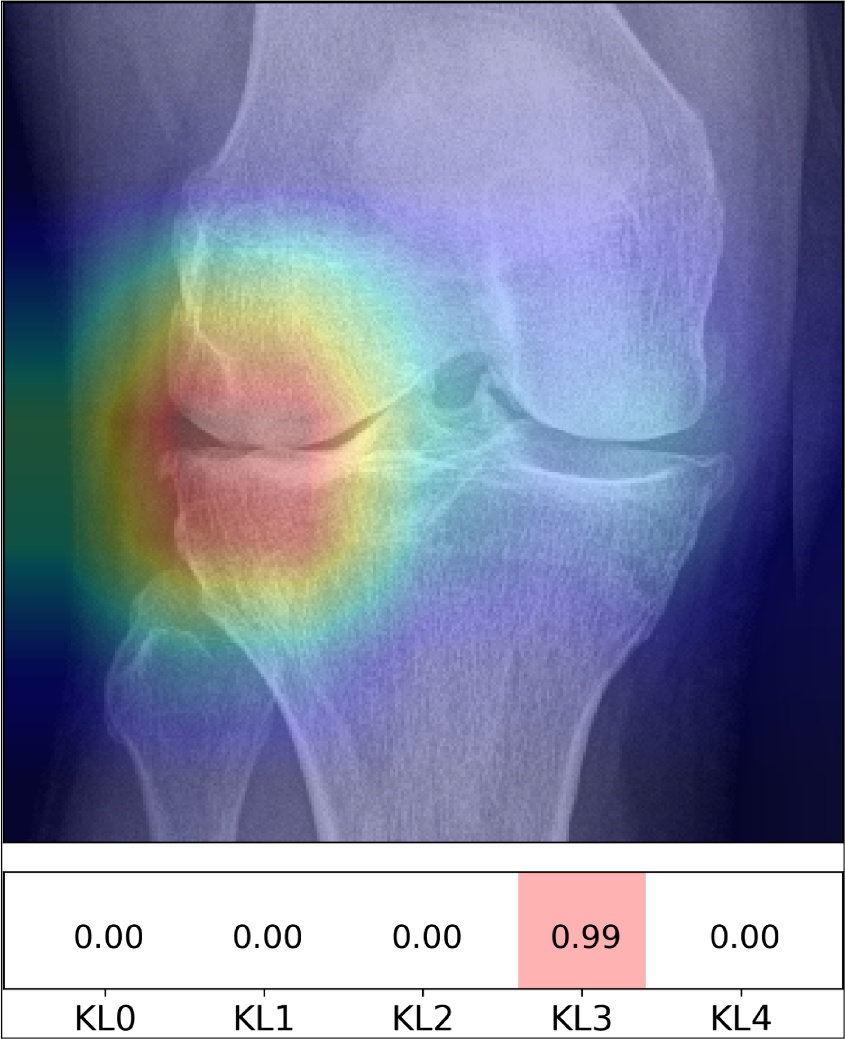}
       }
                \hfill   
     \subfloat[KL-3 -- Our model]{%
       \includegraphics[width=0.29\textwidth,valign=t]{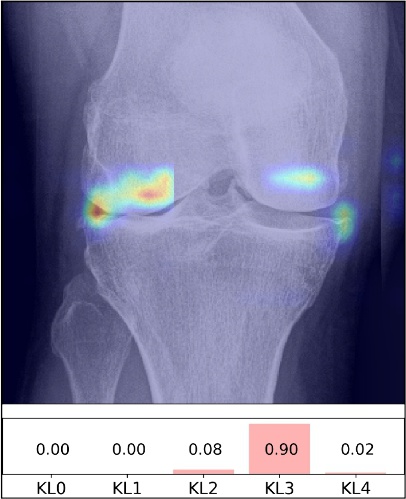}
     }     
     
     \caption{Comparison of the attention maps and output probability distributions between the baseline and our method for detection of moderate osteoarthritis (KL-3).}\label{fig:attention_app2}
\end{figure}

In this section, we present examples of the attention maps produced by the fine-tuned ResNet-34 and our model for clinically relevant cases KL-2 (Figure  \ref{fig:attention_app1}) and also for already present, moderate  OA (Figure  \ref{fig:attention_app2}). The attention maps indicate the benefit of constraining the attention of the network by using prior anatomical knowledge.

\end{document}